\documentclass[letterpaper, 10 pt, journal, twoside]{IEEEtran}
\usepackage{amsmath,amsfonts}
\usepackage{algorithmic}
\usepackage{algorithm}
\usepackage{array}
\usepackage[caption=false,font=normalsize,labelfont=sf,textfont=sf]{subfig}
\usepackage{textcomp}
\usepackage{stfloats}
\usepackage{url}
\usepackage{verbatim}
\usepackage{graphicx}
\usepackage{cite}
\usepackage{hyperref}
\usepackage{color}
\usepackage{makecell}
\begin{document}

\title{RENet: Fault-Tolerant Motion Control for Quadruped Robots via Redundant Estimator Networks under Visual Collapse} 

% Make room for more info lines in the \author command 
\author{Yueqi Zhang$^{\ast1,}$, Quancheng Qian$^{\ast,1}$, Taixian Hou$^{1}$, Peng Zhai$^{\dagger,1}$, Xiaoyi Wei$^{1}$, Kangmai Hu$^{1}$, Jiafu Yi$^{2}$ 

and Lihua Zhang$^{\dagger,1}$%
\thanks{Manuscript received: May, 15, 2025; Revised July, 26, 2025; Accepted August, 30, 2025.}%Use only for final RAL version
\thanks{This paper was recommended for publication by Editor Abderrahmane Kheddar upon evaluation of the Associate Editor and Reviewers' comments.} %Use only for final RAL version
\thanks{$^{1}$Authors are with College of Intelligent Robotics and Advanced Manufacturing, Fudan University, China. (email: {\tt\footnotesize \{zhangyq23, qcqian24, txhou21\}@m.fudan.edu.cn, \{lihuazhang, pzhai\}@fudan.edu.cn}).}%
\thanks{$^{2}$Author is with the School of Information and Communication Engineering, Hainan University, China (email: {\tt\footnotesize yjf1201@163.com})}%
\thanks{$\ast$ Equal Contribution}
\thanks{$\dagger$ Corresponding Author}
% \thanks{Digital Object Identifier (DOI): see top of this page.}
}
% Use only for final RAL version.

% Paper headers
\markboth{IEEE Robotics and Automation Letters. Preprint Version. Accepted september, 2025}
{zhang \MakeLowercase{\textit{et al.}}: RENet: Fault-Tolerant Motion Control via Redundant Estimator Networks under Visual Collapse} 
% Use only for final RAL version


\maketitle

\begin{abstract}
Vision-based locomotion in outdoor environments presents significant challenges for quadruped robots. Accurate environmental prediction and effective handling of depth sensor noise during real-world deployment remain difficult, severely restricting the outdoor applications of such algorithms. To address these deployment challenges in vision-based motion control, this letter proposes the Redundant Estimator Network (RENet) framework. The framework employs a dual-estimator architecture that ensures robust motion performance while maintaining deployment stability during onboard vision failures. Through an online estimator adaptation, our method enables seamless transitions between estimation modules when handling visual perception uncertainties. Experimental validation on a real-world robot demonstrates the framework's effectiveness in complex outdoor environments, showing particular advantages in scenarios with degraded visual perception. This framework demonstrates its potential as a practical solution for reliable robotic deployment in challenging field conditions. Project website: \href{https://RENet-Loco.github.io/}{https://RENet-Loco.github.io/}
\end{abstract}


\section{Introduction}
\IEEEPARstart{I}{n} the field of legged robot control, the state estimator plays a crucial role in environmental perception \cite{fankhauser2018probabilistic} and maintaining dynamic balance \cite{mit-cheetah}. Learning-based implicit estimators, which are trained via supervised learning approaches, are also widely adopted in robust robot control systems \cite{DreamWaQ,RMA}. Existing estimator architectures can be broadly categorized into the "vision + proprioception"\cite{cerberus,optistate} (referred to as \textbf{VP}) and only-proprioception\cite{RMA,DreamWaQ} (referred to as \textbf{OP}) types. VP estimators provide more accurate terrain prediction capabilities and have demonstrated excellent performance in challenging parkour tasks \cite{parkour,exparkour}. However, they remain susceptible to depth sensor noise in real-world deployment environments, which can lead to estimation failures and controller malfunctions. In contrast, OP estimators \cite{DreamWaQ,HIM,SLR} rely exclusively on proprioceptive data, thereby offering stronger generalization across diverse scenarios. Nevertheless, they often struggle in environments with highly varied elevation, such as those involving gap jumping or high-platform climbing.
\begin{figure}[t]
\vspace{0.0cm}
\centerline{\includegraphics[width=9cm]{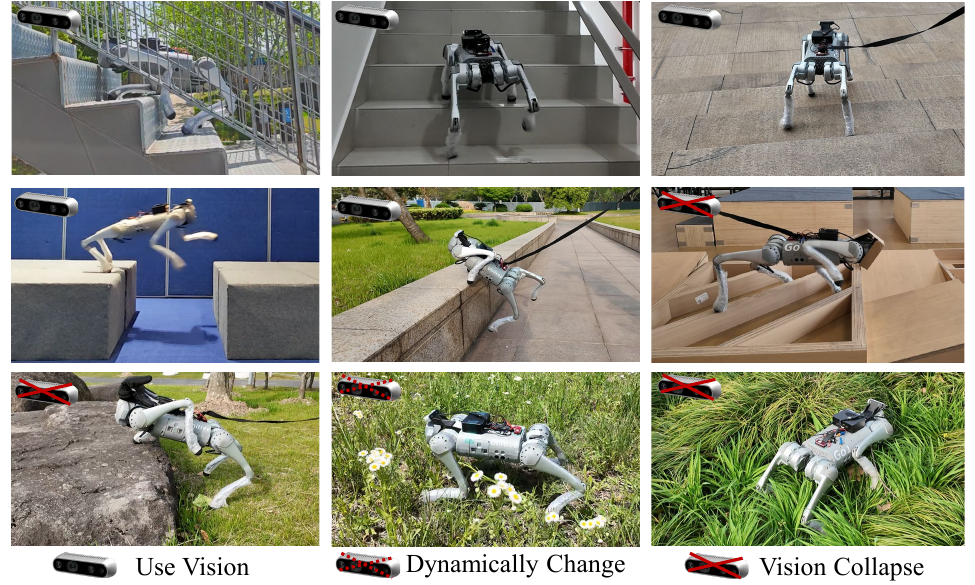}}
\captionsetup{font=scriptsize}
\vspace{-0.2cm}
\caption{Deployment of RENet in diverse environments. Our proposed framework demonstrates robust performance across various outdoor scenarios while maintaining stable operation under severe visual interference.}
\vspace{-0.6cm}
\label{RENet in diverse environments}
\end{figure}
To effectively address these challenges, there is a pressing need for additional research to explore how to better leverage the strengths of both estimators, thus improving the robustness and reliability of quadruped robots in more outdoor tasks.

In nature, animals have evolved various strategies to maintain stability under visual interference. For example, white-tailed deer move agilely under drastic light changes to avoid predators \cite{white-tailed-deer}. Similarly, humans, when vision is blocked by sand or smog, perceive less of their surroundings and often rely on contact to assist motion. For small quadruped robots, vision is often impaired by occlusion or clutter, leading to noisy images. Sole reliance on vision can easily cause failure and controller crash. Hence, robots should automatically detect sensor failure and swiftly switch modes, using environmental contact to aid movement. For example, when a robot enters dense grass and vision is obscured, it could shift to emphasizing proprioception and ground contact. By recognizing leg force patterns and terrain texture, it may adjust gait and posture to stabilize. Once vision recovers, the robot can resume using visual prediction. This switch between sensing modes improves stability in diverse scenarios, maintaining robust locomotion despite significant visual inaccuracy.
\IEEEpubidadjcol

Inspired by the previous discussions, this letter proposes the \textbf{Redundant Estimator Network (RENet)} — a one-stage learning framework with a dual-estimator architecture for quadruped locomotion. The redundant design improves fault tolerance during visual sensor failure. During training, the framework enables smooth switching between estimator inputs. An autonomous estimator selection module is introduced to detect visual noise and trigger timely switching in real-world deployment. To our knowledge, this is the first one-stage vision-fault-tolerant framework with parallel training for legged robots. It retains strong vision-based locomotion while ensuring robustness under high visual uncertainty. The method is deployed on a Unitree GO1 robot, validating its robustness and agility. Our contributions are as follows:
\begin{enumerate}
\item We propose a novel single-stage training framework that enables robust, fault-tolerant quadruped locomotion through fused proprioception and vision perception, even under visual degradation, with online estimator adaptation allowing seamless switching between multiple sensory inputs.
\item We propose a self-adaptive estimator switching mechanism. It enables robots to independently assess noise interference of the depth sensor during deployment and automatically switch between two estimators according to the intensity of environmental noise.
\item We extensively deploy this algorithm across diverse outdoor complex environments, demonstrating its direct real-world transfer capability without fine-tuning.
\end{enumerate}

\section{Related Work}
\subsection{Proprioception-based Locomotion in Legged Robot}
Increasing research focuses on enhancing the robustness of controllers using only robots' proprioception information. Kumar et al. \cite{RMA} propose using 1D-CNN for predicting environmental privileged information, demonstrating that reinforcement learning algorithms can achieve greater robustness in quadruped control than traditional control methods. Nahrendra et al. \cite{DreamWaQ} propose DreamWaQ, which improves the state estimation accuracy of quadrupedal robots by using Context-Aware Inference Networks (CENet). Long et al. \cite{HIM} models environmental disturbances as feedback signals, and enhances the controller's anti-disturbance capability via feedforward compensation. Chen et al. \cite{SLR} uses a self-learning method to train the prediction module without using any privileged information. 

Some recent works have attempted to employ constrained techniques to further enhance sim-to-real stability and simplify the reward function design process. Chane-Sane et al. \cite{CAT} uses early-terminated RL algorithms to ensure stable exploration. Kim et al. \cite{kim2024not} enhances the IPO \cite{IPO} method by introducing a dynamic feasible domain adjustment approach, which simplifies the design of the reward function. Lee et al. \cite{NP3O} conduct a comparative study of different constrained policy optimization algorithms and propose N-P3O algorithm to wheeled-legged robot control tasks.

\subsection{Terrain Reconstruction and Vision-Guided Locomotion}

While blind policies generalize well in many scenarios, they struggle on complex terrains like wide ditches and high platforms, leading to increased research on vision-based locomotion. Fankhauser et al. \cite{fankhauser2018probabilistic, miki2022elevation} reconstruct elevation maps using depth cameras or LiDAR, applicable as a SLAM module in hierarchical systems. Zhuang et al. \cite{parkour} propose a multistage parkour learning framework that distills multiple skills into a unified model. Cheng et al. \cite{exparkour} design a two-stage framework incorporating direction distillation and goal-guided rewards. David et al. \cite{ANYmalparkour} combine pre-trained skills hierarchically for parkour. Miki et al. \cite{miki2022learning} fuse proprioceptive and exteroceptive inputs to handle significant sensor noise. Alexander et al. \cite{optistate} and Yang et al. \cite{cerberus} explore how to integrate legged force feedback to enhance visual state estimation performance.

Two-stage teacher-student distillation often suffers from information loss and performance degradation due to reward function absence and privileged information deficiency. Consequently, recent studies explore one-stage end-to-end alternatives. Luo et al. \cite{PIE} leverage NVIDIA Warp to accelerate depth rendering via CUDA parallelization, significantly reducing memory requirements and replacing complex distillation pipelines. Yu et al. \cite{Sparse} further validate this approach on sparse foothold terrains. However, existing methods neglect visual failure scenarios.

Recently, some works similar to ours have begun addressing this issue. Liu et al. \cite{MBC} use multi-agent reinforcement learning to combine vision and blind policy. Ren et al. \cite{VB-com} achieves policy switching by estimating the cumulative returns of both the blind policy and the perception-based policy. However, their methods still rely on LiDAR-based map reconstruction. This hierarchical deployment leads to error accumulation, which seriously restricts their agility. Meanwhile,  Ren et al. \cite{VB-com} requires additional parameter tuning and complex threshold design to ensure a switch to a blind policy when the visual sensors are subject to noise. In comparison to \cite{MBC}, \cite{VB-com}, our switching mechanism is more flexible and concise. At the same time, their methods lack extensive and diverse experiments in outdoor scenarios.

\section{Method}
In this section, we will introduce the technical details of RENet. In the first subsection, we describe the overall training pipeline. In the second subsection, we discuss how to implement the switching of estimators in real environments. Finally, we introduce some key training techniques.

\subsection{Overview}

\begin{figure*}[htbp]
\vspace{0.0cm}
\centerline{\includegraphics[width=\textwidth]{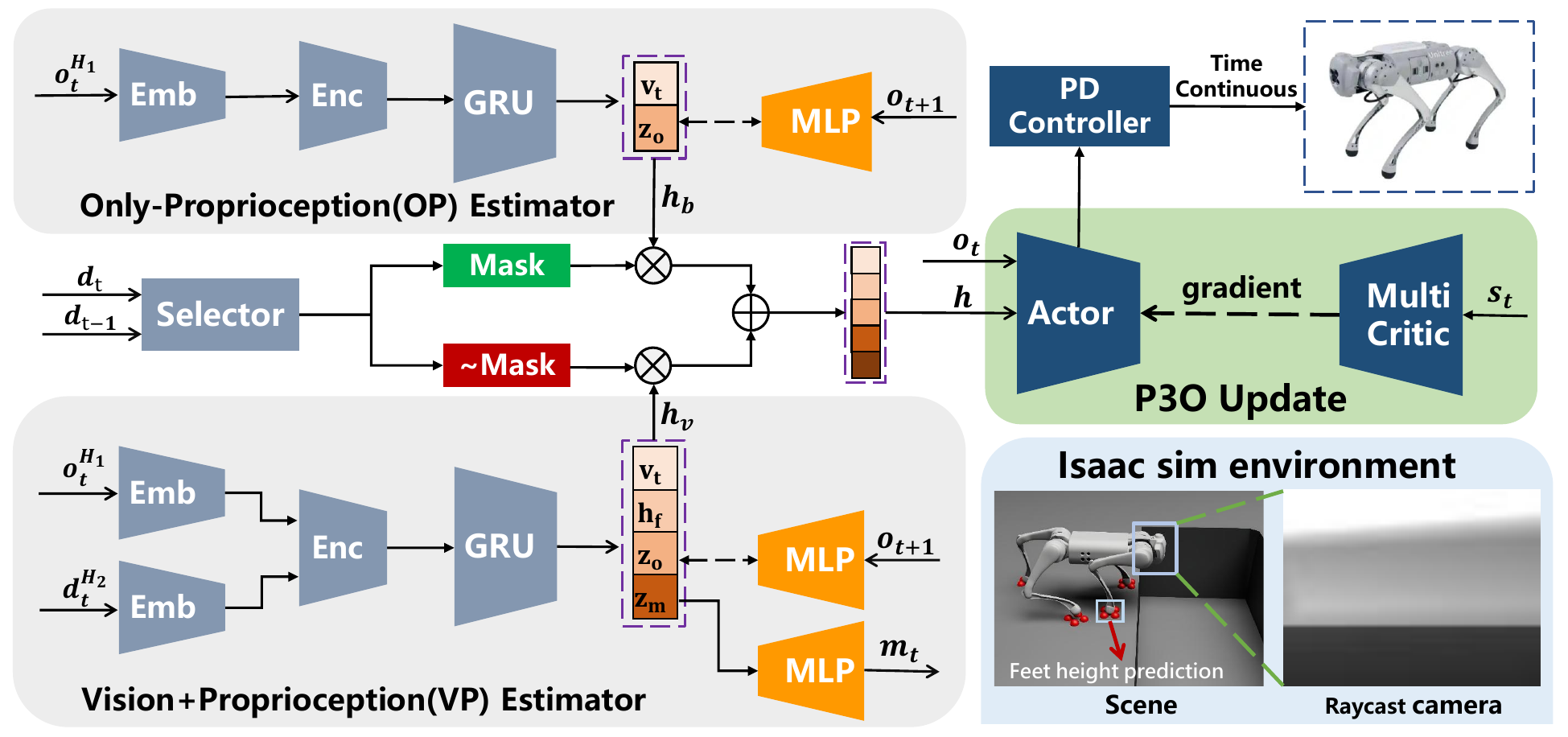}}
\captionsetup{font=scriptsize}
\vspace{-0.3cm}
\caption{Overview of our framework. \textbf{\textcolor[RGB]{150,150,150}{The gray box}} represents the estimator module, running at 10Hz. \textbf{\textcolor[RGB]{84,130,53}{The green box}} indicates the P3O actor-critic architecture, running at 50Hz. \textbf{\textcolor[RGB]{126,195,225}{The blue box}} denotes our simulation environment. We employ raycasting to rapidly generate depth images for policy training. \textbf{\textcolor[RGB]{132,151,176}{Emb}} denotes the embedding module, which compresses the input data into low-dimensional latent variables. \textbf{\textcolor[RGB]{132,151,176}{Enc}} represents the encoder module, responsible for integrating features from data collected at different time steps. \textbf{\textcolor[RGB]{132,151,176}{GRU}} refers to the Gated Recurrent Unit, a type of recurrent neural network used for fusing historical information features. \textbf{\textcolor[RGB]{255,153, 0}{The orange MLP}} represents either the decoder or the target encoder of the HIM \cite{HIM} module, both are not utilized during deployment.}
\vspace{-0.5cm}
\label{pipeline}
\end{figure*}

RENet is an end-to-end one-stage framework that employs NVIDIA Warp's GPU raycaster and CUDA parallelization to efficiently generate depth data across thousands of environments with low memory, the framework is shown in Fig. \ref{pipeline}.

\textbf{Actor-Critic: } To ensure sim-to-real feasibility, we model training as a constrained MDP with limits on joint velocity, torque, and position. The cMDP is solved using NP3O \cite{NP3O} with multiple critics approximating value functions for both rewards and costs. Our \footnotemark{reward} functions remain simple and scenario-agnostic.

\footnotetext[1]{Some key reward functions and their corresponding scales: linear velocity tracking: 1.5 (see Section \ref{training tricks}), angular velocity tracking: 0.5, collision penalty: -10, joint energy penalty: $-1e^{-5}$, action rate penalty: -0.1, default position bias: -0.04, hip position bias: -0.5, joint acceleration penalty: $-2.5e^{-7}$, orientation penalty: -1.}

The Actor network outputs target joint positions, which are converted to torque via an implicit PD controller in Isaac Sim\cite{Isaaclab} operating in near-continuous time.

\textbf{Estimator Networks Training:} As shown in Fig \ref{pipeline}, we define two separate estimators that are trained jointly with the low-level policy. $o_t^{H_1}$ denotes the robot's proprioception data, with a buffer length of $H_1$, and $d_t^{H_2}$ denotes the depth images, with a buffer length of $H_2$. During training, we set the depth image size to $48\times64$ to accelerate training and reduce GPU memory usage, and we also match camera intrinsics matrix between simulation and the real robot to improve simulation-to-real transfer. In our implementation, we set $H_1=10$ and $H_2=2$, because we find that incorporating observations from multiple time steps as input contributes to optimizing the training results. First, all inputs are compressed through an embedding operation. For $o_t^{H_1}$, the embedding is made up of a two-layer MLP network. For the VP estimator, it receives an additional image input $d_t^{H_2}$, which is embedded via a lightweight three-layer CNN. After embedding, the vectors are fed into an encoder module for feature fusion and further dimensionality reduction. The fused vectors are then passed through a one-layer GRU module (64 dimensions) to integrate historical memory information. The final output serves as the latent input for the low-level policy network. For the implementation of the encoder module, we provide two optional schemes:

\begin{itemize}
    \item \textbf{MLP-based feature fusion}: The compressed vectors after embedding are concatenated and then uniformly compressed through an MLP network.
    \item \textbf{Transformer-based \cite{Attention} feature fusion}: The embedded vectors are split into tokens and fused via attention. To ensure fast deployment and training, we implement a simple single-layer single-head attention mechanism.
\end{itemize}

Both independent estimators are trained with supervised learning. To enhance real-world robustness and mitigate network overfitting in the late stage of training, we supervise estimators using the Hybrid Internal Model (HIM), which proves to improve robustness in quadruped robots \cite{HIM}.

For the OP estimator, it is responsible for outputting a predicted linear velocity $v_t$ and a predicted latent variable $z_o$. Therefore, the final loss for the OP estimator is defined as:
\begin{gather}
Loss_{OP} = MSE(v_t,\hat{v_t}) + Loss_{HIM}(z_o, \hat{z}_o),
\end{gather}

where $\hat{z}_o$ denotes the desired latent variable output by the HIM target encoder and $\hat{v}_t$ denotes the ground truth of linear velocity which is obtained from simulator. Similar to the OP estimator, for the VP estimator, in addition to predicting velocity and HIM latent variables, its output also includes the predicted feet height $h_f$ and the predicted 2.5D map $m_t$.
\begin{figure*}[htbp]
\vspace{0.1cm}
\centerline{\includegraphics[width=\textwidth]{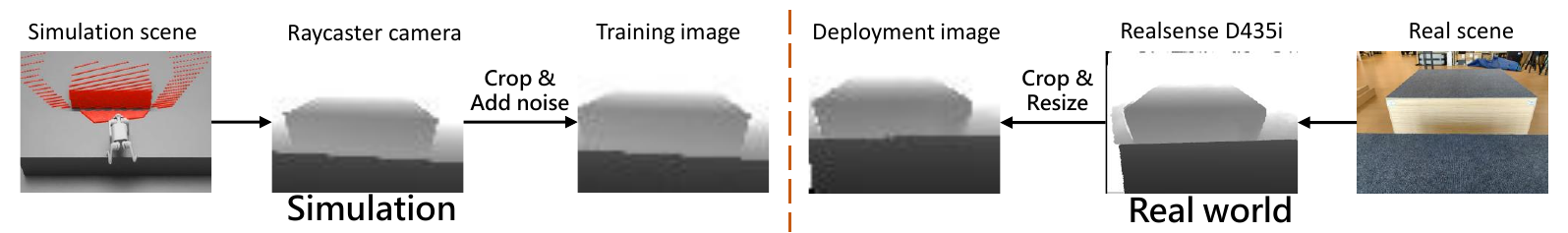}}
\captionsetup{font=scriptsize}
\vspace{-0.4cm}
\caption{The left part shows how to align raycaster outputs with real-world data. The right part illustrates post-processing procedures for real image data.}
\label{raycaster_dr}
\vspace{-0.5cm}
\label{fig3}
\end{figure*}
The feet height prediction as shown in Fig \ref{pipeline} is defined as the average height of the robot's feet to the ground within a localized area (5 cm × 5 cm). We find that predicting feet heights significantly enhances success rates when traversing challenging terrains such as gaps and high platforms. Therefore, the final loss for the VP estimator is defined as:
\begin{gather}
\begin{aligned}
 Loss_{VP} = MSE(v_t,\hat{v_t}) + Loss_{HIM}(z_o, \hat{z}_o) \\+ MSE(h_f,\hat{h_f}) + MSE(m_t,\hat{m_t}).
\end{aligned}
\end{gather}

We propose an \textbf{online estimator adaptation} mechanism within a unified training framework, enabling smooth switching between estimators. Environment sets $S_{op}$ (for OP estimator training) and $S_{vp}$ (for VP estimator training) define their respective training domains. Scenarios are classified by visual dependency: \textbf{simple terrains} (e.g., stairs, uneven ground) where the task can be completed by both estimators, and \textbf{difficult terrains} (e.g., gaps, platforms) requiring visual inputs due to fundamental limitations of proprioception. During training, our adaptation method alternates between different sensory modalities every 20 iterations in simple terrains. When confronting difficult scenarios where visual prediction is essential, we force the use of visual input to guarantee successful task completion.

The final input latent $h$ combines the outputs of both estimators: the VP estimator's $h_v$ and the OP estimator's $h_b$. When the environment selects VP estimator for training, we set $h_b = 0$; when the OP estimator is selected, we correspondingly set $h_v=0$. In our implementation, we define a variable $mask_i$ for each individual training environment to select the the inputs of the estimator. Therefore the expression for the latent variable $h$ can be defined as:
\begin{equation}
\begin{gathered}
h = \text{concat}(h_b * \text{mask}_i, h_v * (1-\text{mask}_i)). \\
\text{mask}_i = 
\begin{cases} 
\mathbf{1}, & i \in S_{op}\\ 
\mathbf{0}, & i \in S_{vp} 
\end{cases}
\end{gathered}
\end{equation}

\subsection{Auto Selector Module}
We design RENet to autonomously identify noise levels and select the OP estimator as input when the VP estimator fails. We assume that the VP estimator is trusted only when the robot deems there is sufficient confidence to rely on visual predictions for the current scene. Otherwise, particularly when encountering severe visual noise or facing complex scenarios beyond the training distribution, the estimator will be deemed unreliable. To achieve this functionality, we jointly train an image-based anomaly detection network during the P3O training process, which progressively encodes a latent space of possible training images as policy converges. The network structure is a CNN-based autoencoder with convolutional and deconvolutional layers. The input to this model consists of depth images of two consecutive frames ($d_t,d_{t-1}$), and the output is the reconstructed depth images ($\hat{d_t},\hat{d}_{t-1}$). The loss function of this module is defined as follows.
\begin{gather}
Loss_{ad}=MSE((d_t,d_{t-1}),(\hat{d_t},\hat{d_{t-1}})).
\end{gather}

When the autoencoder loss exceeds a threshold, indicating high visual sensor unreliability due to extreme noise or unknown simulation environments, the system need autonomously switch to OP estimator mode to maintain stability.

In our experiment, we find that solely relying on the loss to determine when to switch could cause excessive switching frequency, resulting in unnecessary jitter and affecting the smoothness of movement. To mitigate this issue, we define a low-pass filter to limit the frequency of switching.
\begin{gather}
P_{t} = (1-\gamma)P_{t-1}+\gamma \hat{P}_{t}.
\end{gather}

$P_{t}$ indicates the probability of selecting the VP estimator at time $t$. $\gamma$ is a hyperparameter, whose value is not very sensitive. In our experiments, we set $\gamma = 0.1$. If $Loss_{ad}$ $<$ $ \beta$, $\hat{P}_{t}=1$, otherwise $\hat{P}_{t} = 0$. $\beta$ is a hyperparameter; in our tests, this value can be simply set to the maximum value of $Loss_{ad}$ during the simulator testing. During deployment, we can choose to select the VP estimator when the visual probability $P_t$ is larger than $0.5$.

\subsection{Training Techniques} \label{training tricks}

\begin{figure*}[htbp]
\vspace{0.0cm}
\centerline{\includegraphics[width=\textwidth]{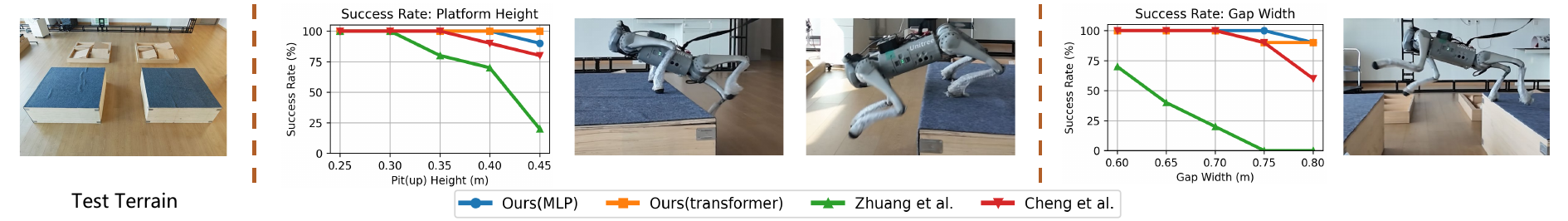}}
\captionsetup{font=scriptsize}
\vspace{-0.38cm}
\caption{Comparison of the parkour performance between our method and other open-source algorithms on the GO1 robot, it demonstrates that our approach achieves the best parkour performance.}
\vspace{-0.6cm}
\label{CP test}
\end{figure*}

\textbf{Raycaster Randomization: } To mitigate the Sim2Real gap from raycaster rendering and avoid policy overfitting, we apply additional processing to the raycaster outputs (see Fig. \ref{raycaster_dr}). The specific methods are as follows:
\begin{itemize}
    \item The camera's depth range is capped at 2 meters. We add 1\% random distance noise and Gaussian noise (variance: 0.01) to raycaster outputs to simulate real depth sensor errors.
    \item To mimic real-world mounting inaccuracies, we apply ±1 cm random positional offsets and ±5° random angular deviations (pitch/yaw) to the raycaster's start pose.
    \item Depth sensors exhibit high-noise at edges. We truncate depth image edges, then resize the valid central region to reduce edge noise interference that causes unstable steering in real applications.
\end{itemize}

\textbf{Zero Velocity Commands Adaptation: } We employ a goal-guided reward for extreme locomotion \cite{exparkour}, but randomly sampling the yaw angles of the target, unlike previous work. However, this method does not maintain stable posture in uneven terrain, compromising robustness and causing control difficulties. Thus, we implement a two-level curriculum: initially training locomotion with commands in [0.2, 1.0], then expanding to [0, 1.0] while adjusting the tracking reward for enhanced stability. The final reward is defined in Eq. \ref{goal-guided reward}.

\begin{equation}
    \label{goal-guided reward}
    \begin{gathered}
    r_{lin\_vel} = 
    \begin{cases} 
    \frac{\min\big(\langle\mathbf{v},\mathbf{n}_{\text{yaw}}\rangle, c_x\big)}{c_x+1e^{-5}}, & c_x \neq 0 \\ 
    \frac{1}{1 + ||\mathbf{v}||}, c_x = 0 
    \end{cases} \\
    \mathbf{n_{yaw}} = (\cos(c_{yaw}),\sin(c_{yaw})).
    \end{gathered}
\end{equation}

Here $c_x$ denotes velocity command, $c_{yaw}$ denotes target yaw relative to the world frame. $\mathbf{v}$ denotes the root velocity relative to the world frame.

\section{EXPERIMENTS}
The experiments in this section are designed to address the following questions:
\begin{itemize}
    \item \textbf{1)} Does each component of RENet contribute to the improvement of the locomotion ability or robustness? What intuitive improvements does it offer compared to a single vision module? (In Section \ref{Exp section 2})
    \item \textbf{2)} Can RENet match or surpass the best performance of existing open-source algorithms on the same quadruped robot platform? Moreover, can it demonstrate agile and robust locomotion across challenging indoor parkour terrains? (In Section \ref{Exp section 3})
    \item  \textbf{3)} Can RENet effectively switch estimators and maintain stable, agile locomotion in complex outdoor environments? (In Section \ref{Exp section 4})
\end{itemize}

\subsection{Experimental Setup and Deployment Details} \label{Exp section 1}
We conduct experiments on the Isaac Sim simulator and the Isaac Lab learning framework \cite{Isaaclab}. RENet achieves robust policy convergence after 15,000 iterations ($\approx$ 8 hours on RTX 4090) using 4,096 parallel environments (about 12GB GPU memory consumption). Training employs terrains with gaps, platforms, stairs, and discrete obstacles under a 10-level progressive curriculum, where higher levels correspond to more difficult terrains. The update mechanism of the curriculum is implemented as described in \cite{walkinminutes}.

Our deployment system employs three ONNX models: a 50Hz low-level action generator, and 10Hz modules for high-level latent representation plus image encoding and reconstruction. The combined size of these models is less than 3 MB ($>10\times$ smaller than \cite{exparkour}), these lightweight models deploy efficiently on low-cost onboard computers without performance degradation.

\subsection{Experiments in Isaac Lab} \label{Exp section 2}

\textbf{Analysis of Comparison Tests:} Experiments are performed to evaluate each module of our method.
\begin{itemize}
    \item Oracle: The policy is trained using the ground truth height map and linear velocity, denoting the upper performance bound in the simulation.
    \item The comparison between MLP and self-attention(SA) in the encoding process.
    \item w/o VEL: This method does not predict the base linear velocity in both estimators.
    \item w/o HIM: This method does not use HIM.
    \item w/o FH: This method does not predict the feet height in the VP estimator.
    \item w/o HF: This method does not predict the height field in the the VP estimator.
    \item One Obs: This method uses a single time-step observation as input.
    \item One Image: This method uses a single image as input to the embedding module.
    \item Zhuang et al. \cite{parkour}: This method distills multiple skills into a single policy.
    \item Cheng et al. \cite{exparkour}: This method enables high-performance parkour by predicting local yaw angles.
\end{itemize}

\setlength{\tabcolsep}{1.6pt}
\renewcommand{\arraystretch}{1.4}
\begin{table}[htbp] 
\vspace{-0.2cm}
\caption{RENET COMPARISON STUDY IN ISAAC SIM}
\vspace{-0.2cm}

\label{Comparison Table}
\begin{center}
\vspace{-0.2cm}
\begin{tabular}{>{\centering\arraybackslash}m{1.4cm}
                |>{\centering\arraybackslash}m{1.1cm}
                >{\centering\arraybackslash}m{1.35cm}
                >{\centering\arraybackslash}m{1.1cm}
                >{\centering\arraybackslash}m{1.35cm}
                >{\centering\arraybackslash}m{1.35cm}}
\Xhline{1.1pt}

Methods     & Gap(↑) & Platform(↑) & Stair(↑) & Collision(↓) & Lin Vel(↑) \\
\hline
Oracle      & 9.96 & 9.97  & 9.93   & 0.026  & 0.5948 \\
Ours(MLP)   & 9.84 & \textbf{9.88} & 9.85 & \textbf{0.070}  & \textbf{0.5598}    \\
Ours(SA)    & \textbf{9.87} & 9.85 & \textbf{9.89} & 0.077  & 0.5556    \\
\hline
w/o VEL     & 7.86 & 4.28 & 6.82 & 0.254  & 0.5221   \\  
w/o FH      & 7.36 & 7.20 & 8.12 & 0.071  & 0.5506  \\
w/o HF      & 5.49 & 5.46 & 6.44 & 0.269  & 0.5155  \\
w/o HIM     & 7.04 & 7.67 & 9.58 & 0.076  & 0.5572  \\
One Obs     & 9.32 & 8.46 & 8.79 & 0.082  & 0.5543  \\
One Image   & 9.63 & 8.86 & 9.82 & 0.104  & 0.5537  \\
\hline
Zhuang \cite{parkour}   & 6.71 & 7.28 & 7.75 & 0.205  & 0.5425  \\
Cheng \cite{exparkour}    & 8.58  & 8.29 & 8.89 & 0.152  & 0.5501  \\
\Xhline{1.1pt}
\end{tabular}
\end{center}
\vspace{-0.5cm}
\end{table}

TABLE \ref{Comparison Table} shows the experimental results, including the highest average curriculum levels in the terrains of gaps, platforms, and stairs, along with collision counts and velocity tracking rewards after 15,000 iterations of training. The pretrained models of Zhuang \cite{parkour} and Cheng\cite{exparkour} are directly transferred to Isaac Lab for evaluation.

\begin{figure*}[htbp]
\centerline{\includegraphics[width=\textwidth]{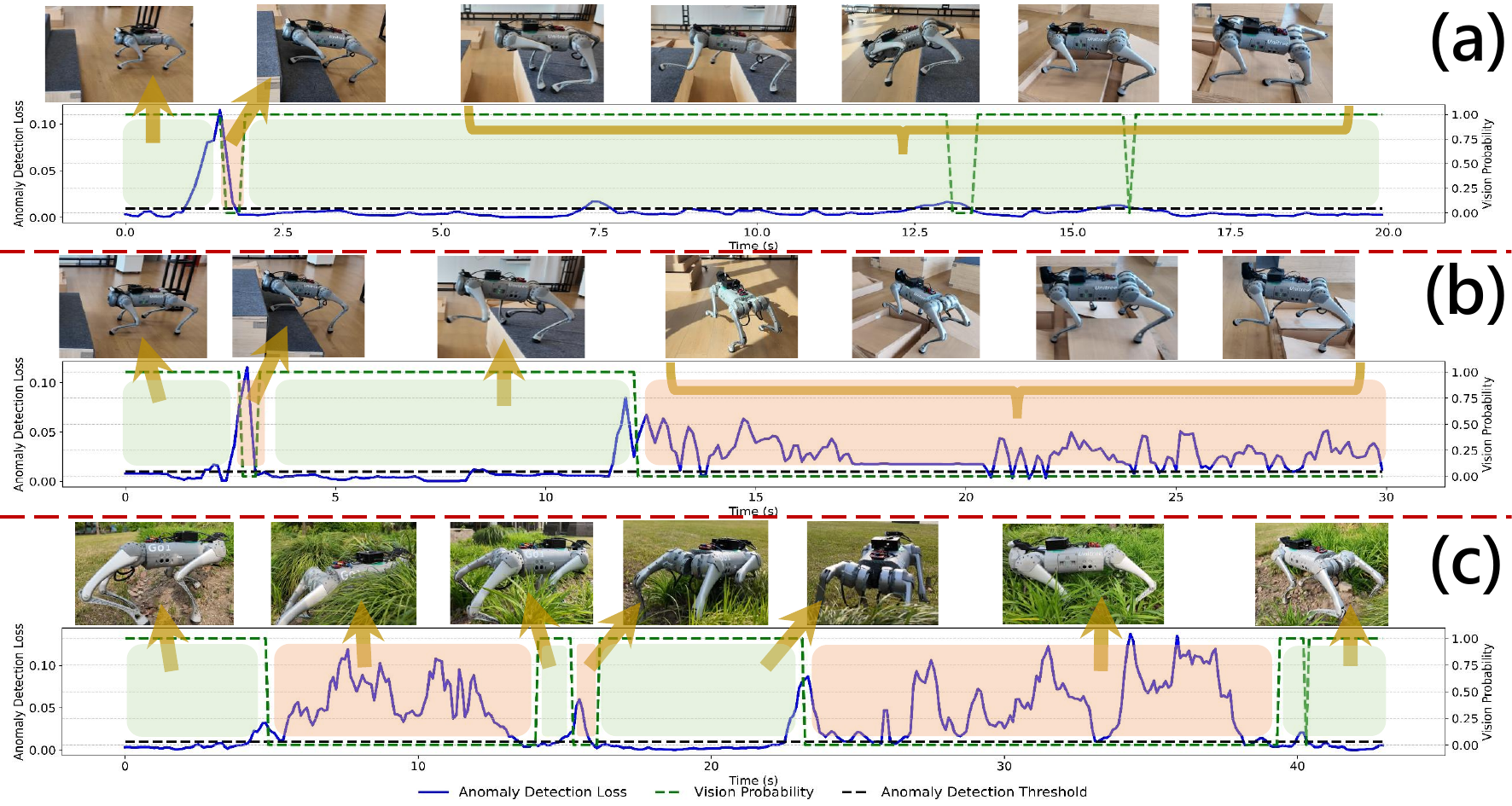}}
\captionsetup{font=scriptsize}
\vspace{-0.4cm}
\caption{Real-world experiments validating RENet's robustness and switching efficacy. 
\textbf{\textcolor[RGB]{238,205,187}{Red regions}}: autonomous switch to OP estimator. 
\textbf{\textcolor[RGB]{190,213,180}{Green regions}}: visual prediction selected. 
\textbf{\textcolor[RGB]{60,60,160}{Blue line}}: estimator's reconstruction loss. 
\textbf{\textcolor[RGB]{100,170,95}{Green dashed line}}: VP estimator probability (\textit{right y-axis}). 
\textbf{Black dashed line}: loss threshold for estimator switching (\textit{left y-axis}). 
All elements reference left y-axis except the \textcolor[RGB]{100,170,95}{green dashed line}.} 
\vspace{-0.6cm}
\label{indoor_exp}
\end{figure*}

The first three rows of TABLE \ref{Comparison Table} show that our method achieves locomotion performance similar to that of the Oracle experiment. In the w/o VEL experiment, the increase in collisions results in a notable decline in motion performance. The robot struggles to learn how to climb higher platforms. In the w/o HIM experiment, due to the lack of state prediction, the robot's ability to traverse gaps and platforms is noticeably affected. In the w/o FH experiment, the robot's performance is affected to some extent across all three terrains. This suggests that foot height helps the robot overcome challenging terrains. In the w/o HF experiment, we observe a significant decline in the robot's traversal performance across all terrains, especially on the gap terrain, where it reached the worst level among several experimental groups. This highlights the critical role of the heightfield in perceiving complex environments. Meanwhile, due to the lack of prediction of the surrounding environment, the robot experiences the highest collision rate and the worst velocity tracking performance. In the One Obs and One Image experiments, the robot exhibits a higher collision rate and reduced agility when traversing the high platform. Furthermore, comparative tests against Zhuang \cite{parkour} and Cheng \cite{exparkour} — both of which employ two-stage distillation — demonstrate RENet's single-stage framework significantly enhances cross-terrain generalization and success rates over traditional multi-skill learning and two-stage distillation algorithms. These results collectively validate rationale and robustness of our method.

Since both MLP and self-attention feature encoder show similar performance. For consistency, we use self-attention in later experiments unless otherwise specified.

\begin{figure}[htbp]
\vspace{-0.2cm}
\centerline{\includegraphics[width=8.9cm]{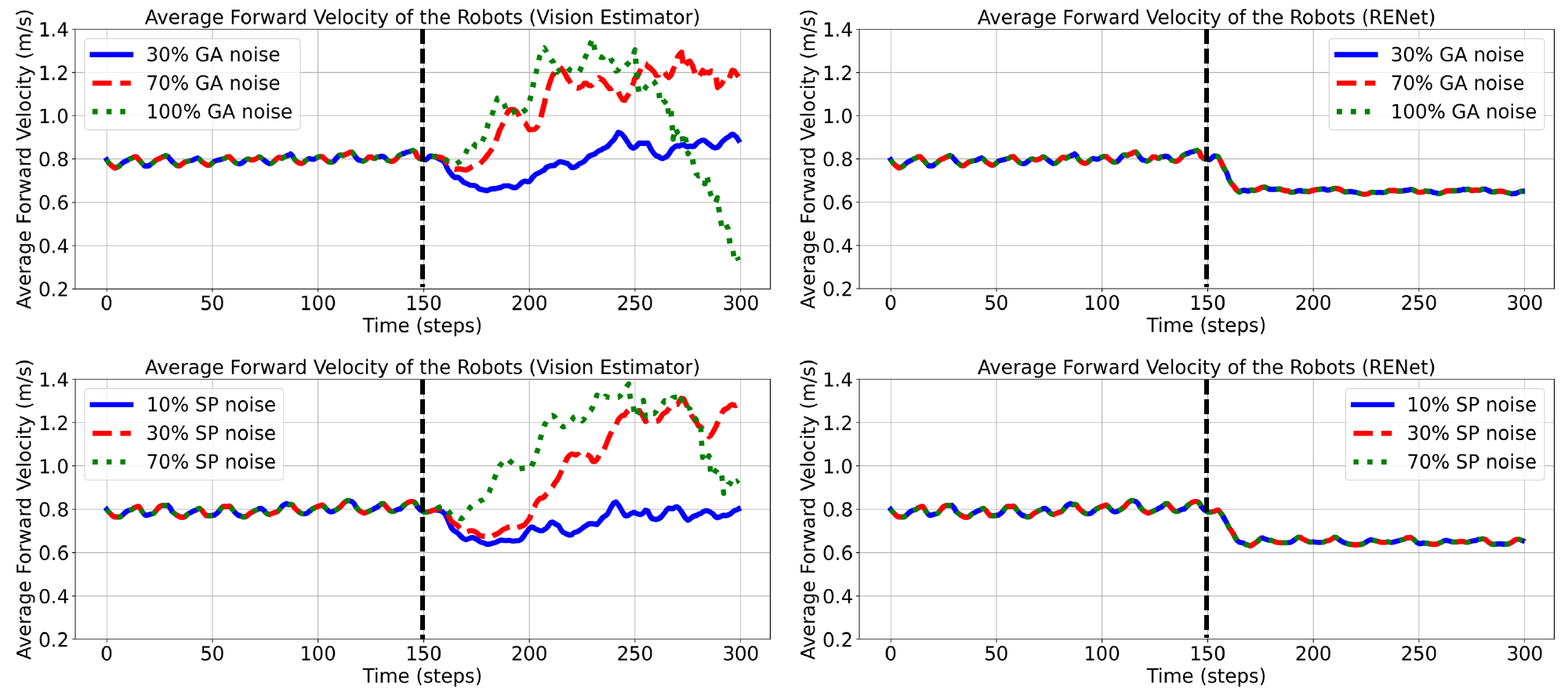}}
\captionsetup{font=scriptsize}
\vspace{-0.3cm}
\caption{Compared to the VP estimator alone, RENet can quickly switch to the OP estimator and maintain stable walking under different visual noise interference, whereas the VP estimator alone struggles to walk properly.}
\vspace{-0.6cm}
\label{single estimator}
\end{figure}

\subsection{Real-World Indoor Experiments} \label{Exp section 3}
\begin{figure*}[htbp]
\vspace{-0.0cm}
\centerline{\includegraphics[width=\textwidth]{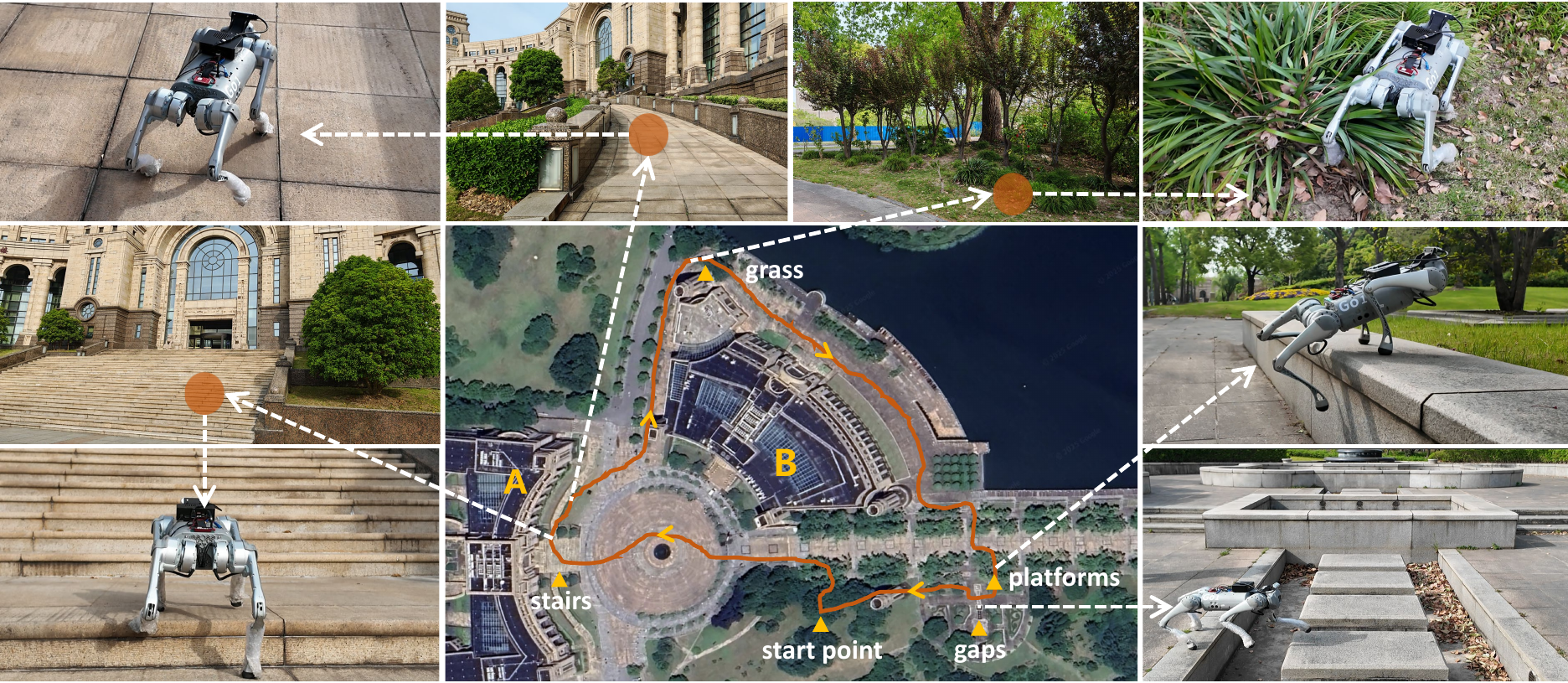}}
\captionsetup{font=scriptsize}
\vspace{-0.4cm}
\caption{ We conduct an approximately 25-minute, 1-kilometer outdoor test with the Unitree GO1 robot. The robot first traverses dense grass, then climbs stairs by building A. Next, it descends a steep slope to reach a grassy area near building B. It finally crosses elevated platforms and gaps between stone blocks before returning to the starting point. }
\vspace{-0.6cm}
\label{longwalk}
\end{figure*}

\textbf{Comparison with Only the VP Estimator: }
In this section, we perform a comparative experiment in Isaac Lab to verify the effectiveness of redundant estimators. As shown in Fig. \ref{single estimator} (a), the horizontal axis represents timesteps, with each timestep equal to 0.02 seconds, and the vertical axis shows the average forward velocity of 20 quadruped robots. They all walk forward on flat ground at a constant command velocity of 0.6 m/s. As Gaussian and salt-and-pepper noise represent prevalent real-world visual sensor degradations, we conduct multiple test groups at timestamp 150 by introducing Gaussian noise (GA noise) at 30\%, 70\%, and 100\% levels, and salt-and-pepper noise (SP noise) at levels 10\%, 30\%, and 70\%, respectively. When using only the VP estimator, these robots exhibit irregular wobbling under 30\% GA noise  or 10\% SP noise. As the noise level increases, they cannot walk properly and even fall (100\% GA noise or 70\% SP noise). In contrast, as shown in Fig. \ref{single estimator} (b), with RENet, all these experiments show that these robots quickly switch to the OP estimator and maintain stable locomotion. Notably, without a penalty for overshooting, these robots consistently move slightly faster than commanded, with either estimator.

\begin{figure}[t]
\centerline{\includegraphics[width=9cm]{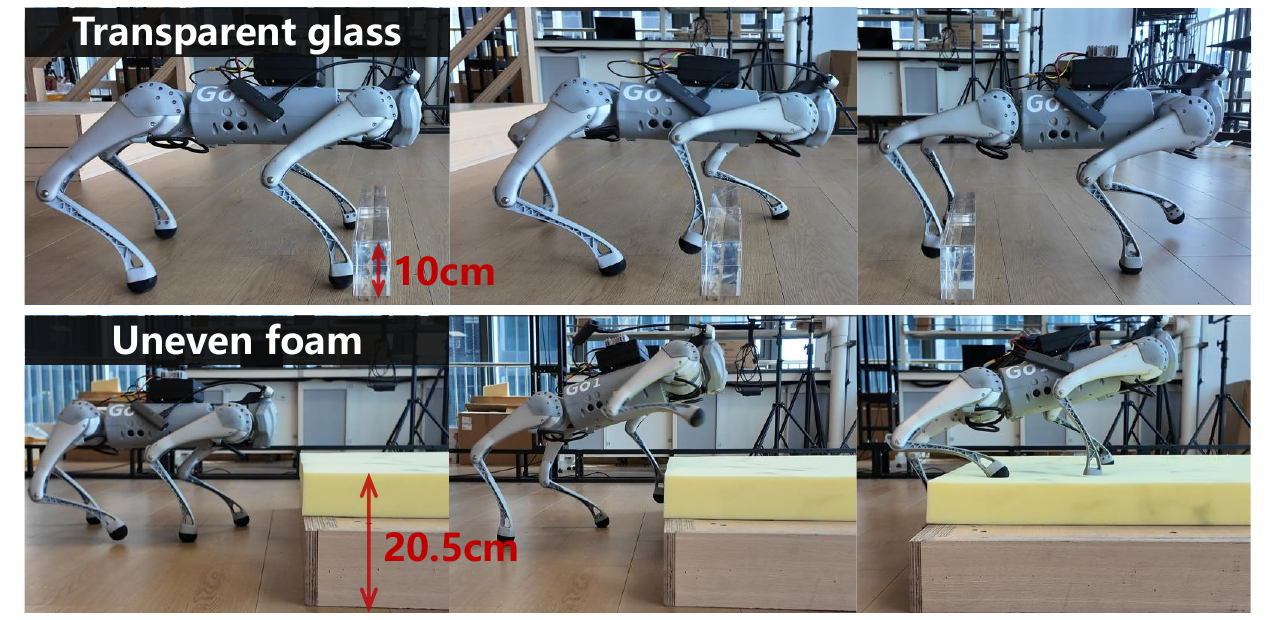}}
    \captionsetup{font=scriptsize}
\vspace{-0.3cm}
\caption{Evaluating the robustness of RENet in visually deceptive scenarios. \textbf{Transparent glass:} Depth sensors fail to measure distances correctly and can only perceive the ground behind. \textbf{Uneven foam:} Heights can be predicted visually, but exhibit significant deviation from true values.}
\label{transparent and foam}
\vspace{-0.5cm}
\end{figure}

Extensive indoor experiments validate RENet's Sim2Real effectiveness, assessing locomotion capability and the robustness of both VP and OP estimators. \textbf{Parkour Ability Tests} (Fig \ref{CP test}) benchmark RENet against \cite{parkour} and \cite{exparkour}. \textbf{Estimator and Selector Tests} (Fig. \ref{indoor_exp} a,b) evaluate performance on split terrains: stair climbing, gap jumping, and platform leaping in the first section, followed by traversal over irregular triangular columns under both normal and camera-occluded conditions. \textbf{Visual Deceptive Tests} (Fig \ref{transparent and foam}) further demonstrate policy adaptation in visually misleading scenarios.

\textbf{Parkour Ability Tests: }
Fig. \ref{CP test} shows Unitree GO1 experiments under identical terrain conditions, repeated 10 times per test. Results compare success rates for traversing specific terrains against \cite{parkour} and \cite{exparkour}, demonstrating RENet's superior locomotion performance. This improvement is attributed to our one-stage training framework, which avoids performance degradation seen in two-stage algorithms.

\begin{figure}[htbp]
\centerline{\includegraphics[width=9cm]{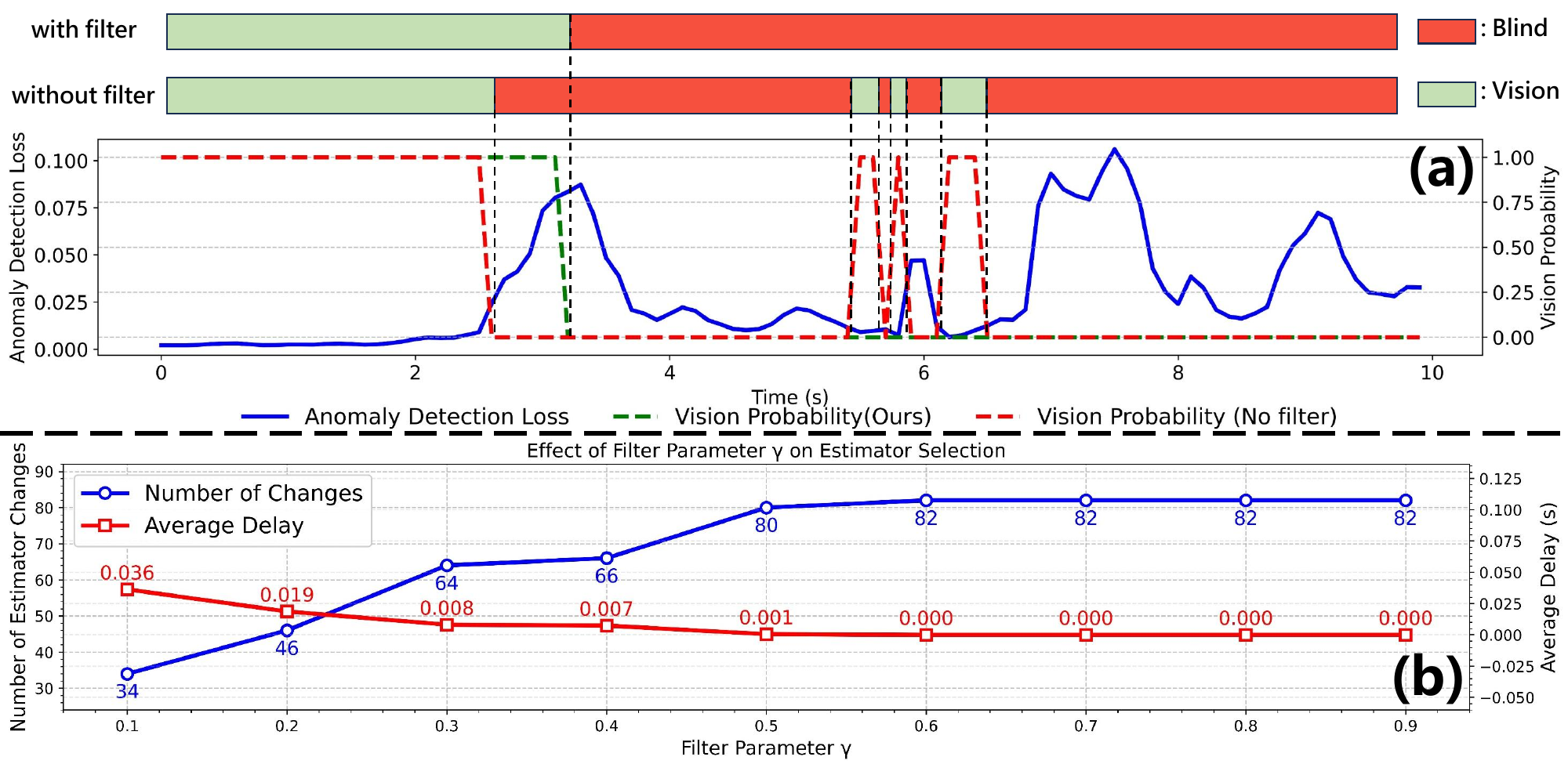}}
    \captionsetup{font=scriptsize}
\vspace{-0.4cm}
\caption{\textbf{Real-world filter comparison}. \textbf{(a)} The filter maintains stability despite slight switching delays, while reducing jitter, limiting switching frequency, and ensuring motion continuity. \textbf{(b)} Varying hyperparameter $\gamma$ affects estimator switching: $\gamma = 0.1$ minimizes unnecessary switches while keeping switching delay below 0.04s.}
\label{comparison with no filter}
\vspace{-0.6cm}
\end{figure}

\textbf{Estimators and Selector Module Tests: } Fig. \ref{indoor_exp} (a) shows RENet’s locomotion ability under full-vision conditions. Interestingly, the selector module consistently switches to the OP estimator when the camera approaches the stair's edge. This is because the RealSense camera cannot measure the accurate depth of obstacles within close proximity (approximately 10 cm). However, the raycaster camera in the simulation does not truncate depth data at close distances. When the robot is near steps, this discrepancy causes the image data to be judged as an anomaly by the automatic switcher, triggering a switch to the OP estimator. However, this transient switch occurs exclusively during continuous high-step climbing and does not compromise overall robustness, while normal-height steps remain unaffected. A potential issue with the selector module is that it might be overly cautious and switch to the OP estimator in any unseen scenario. However, in the latter part of the Fig. \ref{indoor_exp} (a) experiment, when the robot tries to use vision in an untrained environment, the selector doesn’t switch to the OP estimator. This is because the switching threshold is set to the maximum value observed across all successful test environments in Isaac Lab, which prevents the policy from becoming overly conservative. In the latter part of Fig. \ref{indoor_exp} (a), two unexpected switching jitters occur, each lasting under 0.5 seconds. This might be due to environmental interference in the test scenario, yet their short duration doesn’t impact the robot’s movement. Overall, our method shows strong locomotion and stability under full-vision conditions. The experiments demonstrate the selector module isn’t overly conservative in untrained scenarios, showing the VP estimation module’s strong generalization capability.

Fig. \ref{indoor_exp} (b) illustrates estimator switching. Initially, the VP estimator handles challenging terrains. At 12 seconds, camera occlusion triggers a switch to the OP estimator. Despite minor anomaly loss fluctuations from lighting changes, the selector module maintains OP control without reversion, ensuring stable operation.

\textbf{Visual Deceptive Tests: }Fig. \ref{transparent and foam} demonstrates RENet's capability to effectively navigate visually deceptive scenarios. By fusing visual and proprioceptive modalities in our VP estimator, the system autonomously handles partially obscured or misidentified obstacles. Notably, this functionality emerges without task-specific training modifications, indicating the policy's inherent understanding of multi-modal sensory integration for robust terrain adaption.

\subsection{Real-World Outdoor Experiments} \label{Exp section 4}
To further evaluate the effectiveness of the auto selector module, we control the robot to traverse a terrain composed of dense grass, sparse grass, and flat grassy ground, as shown in Fig. \ref{indoor_exp} (c). In flat and sparse grass, the robot uses the VP estimator, while in dense grass, it switches to the OP estimator in time. When moving between grassy and flat terrain repeatedly, the selector module switches estimators according to the level of visual noise. During grass traversal, the anomaly detection occasionally shows sudden loss drops. Fortunately, our filtering mechanism smooths out these disturbances, preventing overly frequent estimator switches. Throughout the process, the robot walks steadily without unexpected behaviors. The outcomes show that our automatic switching mechanism selects the appropriate estimator to take over and run the policy smoothly. Overall, the experiments show that our algorithm achieves robust locomotion performance and good generalization ability in complex outdoor conditions. Furthermore, in Fig. \ref{comparison with no filter} we evaluates the effect of $\gamma$ on switching delay and count, based on combined data from three real-world indoor and outdoor trajectories. Figure \ref{comparison with no filter} (a) demonstrates that adding a low-pass filter effectively reduces unnecessary switching, and Figure \ref{comparison with no filter} (b) further shows that with $\gamma=0.1$, the system achieves rapid mode switching within 0.04 seconds while minimizing superfluous transitions. Real-robot experiments (Fig. \ref{indoor_exp} and Fig \ref{longwalk}) verify that this hyperparameter setting maintains robustness and ensures correct estimator switching in time. 

Fig. \ref{longwalk} shows a long-distance stress test in diverse outdoor environments. The robot successfully traverses dense grass, stairs, slopes, and even untrained scenarios like continuous gaps, including climbing a 48cm platform, all without intervention. These results confirm RENet's robustness and deployability in complex real-world settings.

\section{CONCLUSION}
In this letter, we propose RENet, an innovative one-stage, end-to-end framework for training motion control policies for quadruped robots. Our framework not only enables agile parkour capabilities, but also demonstrates remarkable robustness to severe visual disturbances. Our proposed anomaly detection based selector module facilitates automatic estimator switching for visual fault-tolerant control. This feature extends the application of vision-based motion control algorithms to more complex and challenging environments. The limitation is that the current work lacks full 3D semantic understanding and has limited robustness to lighting variations, especially under strong outdoor lighting conditions. Future work will focus on incorporating advanced semantic perception capabilities, continuously improving the robustness and safety of RL algorithms, and exploring the applicability of this framework on humanoid robots with more intricate structures.

\bibliographystyle{IEEEtran}
\bibliography{mylib}

\begin{thebibliography}{10}
\providecommand{\url}[1]{#1}
\csname url@samestyle\endcsname
\providecommand{\newblock}{\relax}
\providecommand{\bibinfo}[2]{#2}
\providecommand{\BIBentrySTDinterwordspacing}{\spaceskip=0pt\relax}
\providecommand{\BIBentryALTinterwordstretchfactor}{4}
\providecommand{\BIBentryALTinterwordspacing}{\spaceskip=\fontdimen2\font plus
\BIBentryALTinterwordstretchfactor\fontdimen3\font minus \fontdimen4\font\relax}
\providecommand{\BIBforeignlanguage}[2]{{%
\expandafter\ifx\csname l@#1\endcsname\relax
\typeout{** WARNING: IEEEtran.bst: No hyphenation pattern has been}%
\typeout{** loaded for the language `#1'. Using the pattern for}%
\typeout{** the default language instead.}%
\else
\language=\csname l@#1\endcsname
\fi
#2}}
\providecommand{\BIBdecl}{\relax}
\BIBdecl

\bibitem{fankhauser2018probabilistic}
P.~Fankhauser, M.~Bloesch, and M.~Hutter, ``Probabilistic terrain mapping for mobile robots with uncertain localization,'' \emph{IEEE Robotics and Automation Letters}, vol.~3, no.~4, pp. 3019--3026, 2018.

\bibitem{mit-cheetah}
G.~Bledt, M.~J. Powell, B.~Katz, J.~Di~Carlo, P.~M. Wensing, and S.~Kim, ``Mit cheetah 3: Design and control of a robust, dynamic quadruped robot,'' in \emph{2018 IEEE/RSJ International Conference on Intelligent Robots and Systems (IROS)}, 2018, pp. 2245--2252.

\bibitem{DreamWaQ}
I.~M.~A. Nahrendra, B.~Yu, and H.~Myung, ``Dreamwaq: Learning robust quadrupedal locomotion with implicit terrain imagination via deep reinforcement learning,'' in \emph{2023 IEEE International Conference on Robotics and Automation (ICRA)}.\hskip 1em plus 0.5em minus 0.4em\relax IEEE, 2023, pp. 5078--5084.

\bibitem{RMA}
A.~Kumar, Z.~Fu, D.~Pathak, and J.~Malik, ``Rma: Rapid motor adaptation for legged robots,'' \emph{Robotics: Science and Systems XVII}, 2021.

\bibitem{cerberus}
S.~Yang, Z.~Zhang, Z.~Fu, and Z.~Manchester, ``Cerberus: Low-drift visual-inertial-leg odometry for agile locomotion,'' in \emph{2023 IEEE International Conference on Robotics and Automation (ICRA)}, 2023, pp. 4193--4199.

\bibitem{optistate}
A.~Schperberg, Y.~Tanaka, S.~Mowlavi, F.~Xu, B.~Balaji, and D.~Hong, ``Optistate: State estimation of legged robots using gated networks with transformer-based vision and kalman filtering,'' in \emph{2024 IEEE International Conference on Robotics and Automation (ICRA)}, 2024, pp. 6314--6320.

\bibitem{parkour}
Z.~Zhuang, Z.~Fu, J.~Wang, C.~G. Atkeson, S.~Schwertfeger, C.~Finn, and H.~Zhao, ``Robot parkour learning,'' in \emph{Conference on Robot Learning}.\hskip 1em plus 0.5em minus 0.4em\relax PMLR, 2023, pp. 73--92.

\bibitem{exparkour}
X.~Cheng, K.~Shi, A.~Agarwal, and D.~Pathak, ``Extreme parkour with legged robots,'' in \emph{2024 IEEE International Conference on Robotics and Automation (ICRA)}.\hskip 1em plus 0.5em minus 0.4em\relax IEEE, 2024, pp. 11\,443--11\,450.

\bibitem{HIM}
J.~Long, Z.~Wang, Q.~Li, L.~Cao, J.~Gao, and J.~Pang, ``Hybrid internal model: Learning agile legged locomotion with simulated robot response,'' in \emph{The Twelfth International Conference on Learning Representations}, 2024.

\bibitem{SLR}
S.~Chen, Z.~Wan, S.~Yan, C.~Zhang, W.~Zhang, Q.~Li, D.~Zhang, and F.~U.~D. Farrukh, ``Slr: Learning quadruped locomotion without privileged information,'' in \emph{Conference on Robot Learning}.\hskip 1em plus 0.5em minus 0.4em\relax PMLR, 2025, pp. 3212--3224.

\bibitem{white-tailed-deer}
B.~A. Newman, J.~R. Dyal, K.~V. Miller, M.~J. Cherry, and G.~J. D'Angelo, ``Influence of visual perception on movement decisions by an ungulate prey species,'' \emph{Biology Open}, vol.~12, no.~10, 2023.

\bibitem{CAT}
E.~Chane-Sane, P.-A. Leziart, T.~Flayols, O.~Stasse, P.~Sou{\`e}res, and N.~Mansard, ``Cat: Constraints as terminations for legged locomotion reinforcement learning,'' in \emph{2024 IEEE/RSJ International Conference on Intelligent Robots and Systems (IROS)}.\hskip 1em plus 0.5em minus 0.4em\relax IEEE, 2024, pp. 13\,303--13\,310.

\bibitem{kim2024not}
Y.~Kim, H.~Oh, J.~Lee, J.~Choi, G.~Ji, M.~Jung, D.~Youm, and J.~Hwangbo, ``Not only rewards but also constraints: Applications on legged robot locomotion,'' \emph{IEEE Transactions on Robotics}, 2024.

\bibitem{IPO}
Y.~Liu, J.~Ding, and X.~Liu, ``Ipo: Interior-point policy optimization under constraints,'' in \emph{Proceedings of the AAAI conference on artificial intelligence}, vol.~34, no.~04, 2020, pp. 4940--4947.

\bibitem{NP3O}
J.~Lee, L.~Schroth, V.~Klemm, M.~Bjelonic, A.~Reske, and M.~Hutter, ``Evaluation of constrained reinforcement learning algorithms for legged locomotion,'' \emph{arXiv preprint arXiv:2309.15430}, 2023.

\bibitem{miki2022elevation}
T.~Miki, L.~Wellhausen, R.~Grandia, F.~Jenelten, T.~Homberger, and M.~Hutter, ``Elevation mapping for locomotion and navigation using gpu,'' in \emph{2022 IEEE/RSJ International Conference on Intelligent Robots and Systems (IROS)}.\hskip 1em plus 0.5em minus 0.4em\relax IEEE, 2022, pp. 2273--2280.

\bibitem{ANYmalparkour}
\BIBentryALTinterwordspacing
D.~Hoeller, N.~Rudin, D.~Sako, and M.~Hutter, ``Anymal parkour: Learning agile navigation for quadrupedal robots,'' \emph{Science Robotics}, vol.~9, no.~88, p. eadi7566, 2024. [Online]. Available: \url{https://www.science.org/doi/abs/10.1126/scirobotics.adi7566}
\BIBentrySTDinterwordspacing

\bibitem{miki2022learning}
T.~Miki, J.~Lee, J.~Hwangbo, L.~Wellhausen, V.~Koltun, and M.~Hutter, ``Learning robust perceptive locomotion for quadrupedal robots in the wild,'' \emph{Science robotics}, vol.~7, no.~62, p. eabk2822, 2022.

\bibitem{PIE}
S.~Luo, S.~Li, R.~Yu, Z.~Wang, J.~Wu, and Q.~Zhu, ``Pie: Parkour with implicit-explicit learning framework for legged robots,'' \emph{IEEE Robotics and Automation Letters}, 2024.

\bibitem{Sparse}
R.~Yu, Q.~Wang, Y.~Wang, Z.~Wang, J.~Wu, and Q.~Zhu, ``Walking with terrain reconstruction: Learning to traverse risky sparse footholds,'' \emph{arXiv preprint arXiv:2409.15692}, 2024.

\bibitem{MBC}
H.~Liu, Y.~Cheng, R.~Li, X.~Hu, L.~Ye, and H.~Liu, ``Mbc: Multi-brain collaborative control for quadruped robots,'' in \emph{Conference on Robot Learning}.\hskip 1em plus 0.5em minus 0.4em\relax PMLR, 2025, pp. 3688--3704.

\bibitem{VB-com}
J.~Ren, T.~Huang, H.~Wang, Z.~Wang, Q.~Ben, J.~Pang, and P.~Luo, ``Vb-com: Learning vision-blind composite humanoid locomotion against deficient perception,'' \emph{arXiv preprint arXiv:2502.14814}, 2025.

\bibitem{Isaaclab}
M.~Mittal, C.~Yu, Q.~Yu, J.~Liu, N.~Rudin, D.~Hoeller, J.~L. Yuan, R.~Singh, Y.~Guo, H.~Mazhar, A.~Mandlekar, B.~Babich, G.~State, M.~Hutter, and A.~Garg, ``Orbit: A unified simulation framework for interactive robot learning environments,'' \emph{IEEE Robotics and Automation Letters}, vol.~8, no.~6, pp. 3740--3747, 2023.

\bibitem{Attention}
A.~Vaswani, ``Attention is all you need,'' \emph{Advances in Neural Information Processing Systems}, 2017.

\bibitem{walkinminutes}
N.~Rudin, D.~Hoeller, P.~Reist, and M.~Hutter, ``Learning to walk in minutes using massively parallel deep reinforcement learning,'' in \emph{Conference on Robot Learning}.\hskip 1em plus 0.5em minus 0.4em\relax PMLR, 2022, pp. 91--100.

\end{thebibliography}
\vfill
\end{document}